\documentclass{article}

\usepackage[final, nonatbib]{nips_2018}

\usepackage[utf8]{inputenc} 
\usepackage[T1]{fontenc}    
\usepackage{hyperref}       
\usepackage{url}            
\usepackage{booktabs}       
\usepackage{amsfonts}       
\usepackage{nicefrac}       
\usepackage{microtype}      
\usepackage{graphicx}
\usepackage{amsmath}
\usepackage{amsmath}
\DeclareMathOperator*{\argmax}{argmax} 
\usepackage{enumitem}
\setitemize{leftmargin=11pt, parsep=5pt}
\usepackage{titlesec}
\titleformat{\section}{\normalfont\Large\bfseries}{\thesection}{1em}{}
\titleformat{\paragraph}[runin]{\normalfont\normalsize\bfseries}{\theparagraph}{1em}{}
\titlespacing*{\section} {0pt}{0.5ex plus 0.2ex minus .2ex}{0.5ex plus .2ex}
\titlespacing*{\paragraph} {0pt}{0.2em}{1em}

\usepackage{caption}
\usepackage{subcaption}

\title{FoldingZero: Protein Folding from Scratch in Hydrophobic-Polar Model}

\author{
    Yanjun Li, Hengtong Kang, Ketian Ye, Shuyu Yin and Xiaolin Li\thanks{Corresponding author} \thanks{This work was supported in part by National Science Foundation (CNS-1624782, CNS-1747783), National Institutes of Health (R01-GM110240), and Industrial Members of NSF Center for Big Learning (CBL).}\\
    Large-scale Intelligent Systems Laboratory, NSF Center for Big Learning, University of Florida\\
    \texttt{\{yanjun.li, hengtongkang, k.ye, shuyuyin\}@ufl.edu, andyli@ece.ufl.edu}
}

\begin{document}

\maketitle

\begin{abstract}
  \textit{De novo} protein structure prediction from amino acid sequence is one of the most challenging problems in computational biology. As one of the extensively explored mathematical models for protein folding, Hydrophobic-Polar (HP) model enables thorough investigation of protein structure formation and evolution. Although HP model discretizes the conformational space and simplifies the folding energy function, it has been proven to be an NP-complete problem. In this paper, we propose a novel protein folding framework FoldingZero, self-folding a \textit{de novo} protein 2D HP structure from scratch based on deep reinforcement learning. FoldingZero features the coupled approach of a two-head (policy and value heads) deep convolutional neural network (HPNet) and a regularized Upper Confidence Bounds for Trees (R-UCT). It is trained solely by a reinforcement learning algorithm, which improves HPNet and R-UCT iteratively through iterative policy optimization. Without any supervision and domain knowledge, FoldingZero not only achieves comparable results, but also learns the latent folding knowledge to stabilize the structure. Without exponential computation, FoldingZero shows promising potential to be adopted for real-world protein properties prediction. 
\end{abstract}

\section{Introduction}
Proteins are complex biological macromolecules that play critical roles in the body.  In standard terms, proteins always naturally fold to the same unique 3-dimensional structures, which are known as their native conformation. Based on the thermodynamic hypothesis of Christian Anfinsen~\cite{kresge2006thermodynamic}, the native conformation is only determined by the sequence of amino acid and formed via a physical process named protein folding. How to devise a computer algorithm to predict the protein structures from the sequences is one of the most challenging and fundamental problems in computational biology, molecular biology, and  theoretical chemistry. It attracts lots of research attention for its significant impacts and applications in disease prediction~\cite{sundaram2018predicting}, protein design~\cite{lee2018novo} and so on.

The Hydrophobic-Polar (HP) model proposed by Dill~\cite{dill1985theory,lau1989lattice} is one of the extensively studied mathematical models for protein folding.  In the HP model, 20 different types of amino acids are classified as hydrophobic (H) or polar (P) by the degree of their hydrophobicity. It simplifies the protein sequence based on the fact that the hydrophobic interaction is a significant factor in the folding process. The hydrophobic amino acids are predominantly located in the folded protein’s core because they must have less contact with water, whereas the polar ones are more commonly on the surface \cite{dill2012protein}. The sequence is “folded” as a self-avoiding walk on a 2D or 3D lattice, such that the vertices of the lattice can be occupied by at most one amino acid, and the adjacent amino acids in the protein sequence must also occupy adjacent lattice vertices. 2D square based lattice is usually utilized as a benchmark for evaluating the algorithm. The HP model considers the interaction between two amino acids only if the pairwise residues are closest neighbors on lattices but not adjacent in the chain. It assigns a negative one energy value to a contact between adjacent, non-covalently bound H-H residues, and zero value to P-H and P-P contacts. The target of folding algorithm is to discover the protein native conformation with the lowest energy value, which equals to maximize the number of H-H contacts on the lattice. 

Although the HP model discretizes the conformational space and simplifies the folding energy function, it has been proven as a NP-complete problem \cite{unger1993finding,berger1998protein,crescenzi1998complexity}. Therefore, it is computationally intractable to reach the globally optimal solution in the HP model, especially with the increase of the protein sequence length.  

In this paper, we propose a novel and efficient framework {\bf FoldingZero} to self-fold protein 2D HP structure based on deep reinforcement learning. It's the first folding from scratch solution in this one of the 21st century open grand challenges. This is ultimately needed with transformative impacts because we have a huge amount of sequenced protein data without structure annotations, which are fundamentally critical for protein functions, gene defect, disease detection and remedy.


The key contributions of this work are multifold.

\begin{itemize} [topsep=0pt,itemsep=4pt,partopsep=0pt, parsep=0pt]
\item Within our knowledge, this is the first work that uses deep reinforcement learning technique to solve the challenging protein folding problem. We attempt to usher in the high-impact artificial intelligence tool to empower fundamental life science research.
\item We propose a novel protein folding framework FoldingZero to self-fold the \textit{de novo} protein 2D HP structure from scratch based on the coupled approach of a two-head deep convolutional neural network (HPNet) and a regularized Upper ConfidenceBounds for Trees (R-UCT).
\item Although the folding scenario discussed in the paper focuses on the HP model, the FoldingZero approach can be generalized to more complicated protein models and meet more real-world needs in computational biology and chemistry. 
\item Without any domain knowledge, FoldingZero learns from scratch and eventually achieve the comparable results on the benchmark dataset. It also learns latent folding knowledge to stabilize the protein structure.
\end{itemize}




\begin{figure*}[!b]
\vspace{-1.0em}
\centering
\includegraphics[scale=0.40]{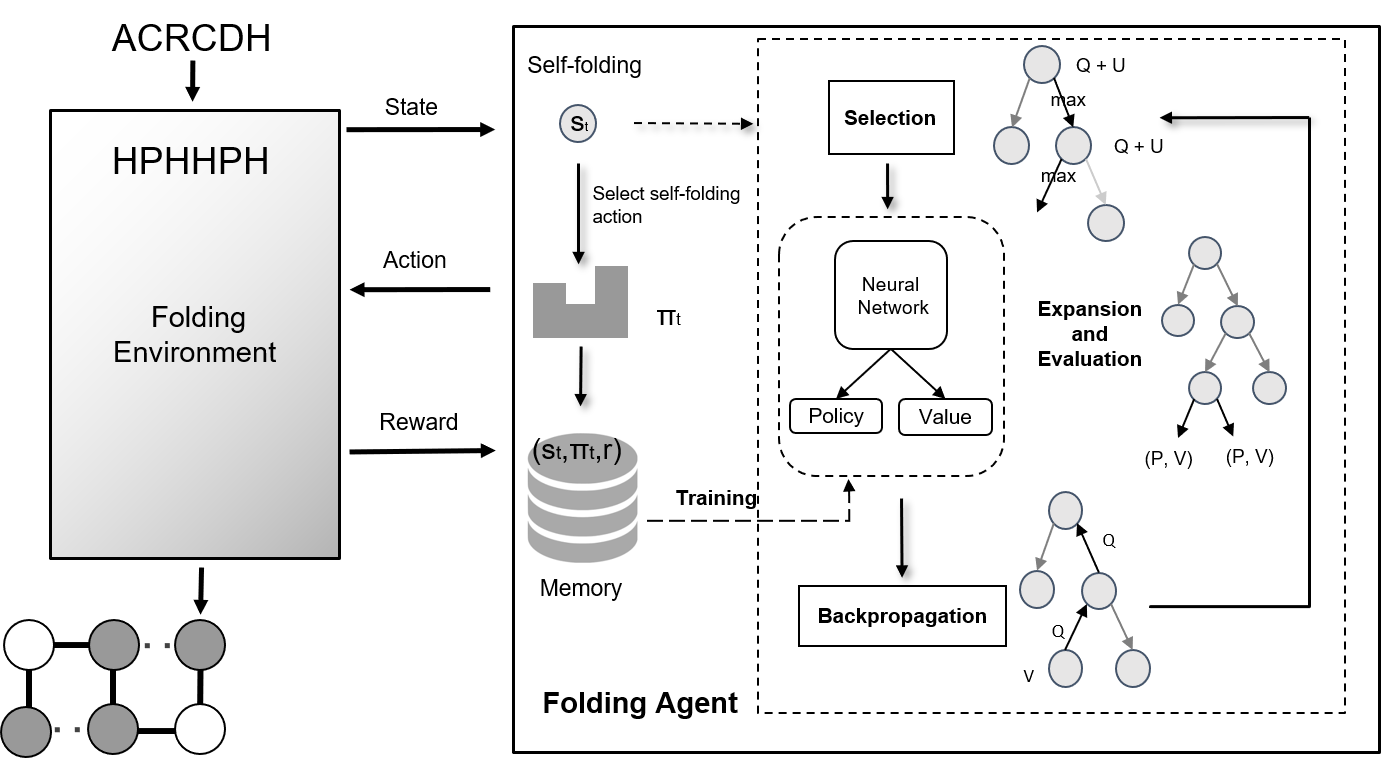}
\vspace{1.0em}
\caption{\textbf{FoldingZero framework} describes the interaction between the environment (left one), and the folding agent (right one). Starting from an initial state given by the environment, the agent carries out simulations, including selection, evaluation, expansion and backpropagation processes. The most promising folding position is selected to self-fold the next residue. When the folding terminates, states, rewards and policies will be stored into the memory to train the HPNet.}
\label{fig_framework}
\end{figure*}

\section{Methods}

The proposed FoldingZero architecture consists of two components: a HP folding environment and a self-folding agent as illustrated in Figure~\ref{fig_framework}. Based on the current folding state given by the environment, the agent sequentially self-folds the amino acid along the protein sequence. For example, the agent randomly places the first amino acid in the environment. Based on the state of the first amino acid, the agent uses its trained model to place the second amino acid next to the first one. This process continues until the agent folds the last amino acid in the sequence. With the self-folding process, the final H-H contact score is given by the environment. The score is utilized as a reward to evaluate each folding action. 


\subsection{HP folding environment}
Protein primary sequence is typically notated as a string of letters. In the environment, each amino acid in the sequence is firstly translated to H or P type according to its chemical properties. For example, given a primary sequence ACRCDH, its HP representation is {HPHHPH}. 

Starting from the first one, each amino acid will be self-folded by the agent on the 2D grid. The environment defines the folding state at time-step $t$ as $s_t$ and corresponding legal action as $a \in \mathcal{A}(s)$. Action space of each state contains at most 3 possible moves (forward, left and right) because of self-avoiding. Only the vertex of the lattice can be occupied. The neighboring residues in the protein sequence must also occupy adjacent vertices. Every folding action leads to a new folding state $s_{t+1}$, which contains all so far folded amino acids’ positions on the 2D lattice. When the folding is done for all the residues along the sequence, the amount of final H-H contact is calculated as $r$, which will be fed back to the agent as self-folding reward.

In FoldingZero, the lattice is represented as a 3D tensor with 2D grid (height and width) and 1D channel (analogous to RGB channels in images). Each grid point in the tensor corresponds to either vertex or edge of the 2D lattice. Vertex can be occupied by two types of amino acids, such as H or P. We also define two connection types on edge. One denotes the “primary connect” between adjacent residues on the primary sequence and the other denotes the ``H-H contact'' between pairwise H residues which are the closest neighbors on lattices but not adjacent on the primary sequence. Thus, 4 binary channels with value 0 or 1 are utilized to represent one grid point; only one channel can be activated, and the others are all zeroes.  
\vspace{-1.0em}
\subsection{Self-folding mechanism}
\vspace{-0.5em}
The agent in FoldingZero incorporates two interactive components, HPNet and R-UCT. HPNet takes the folding state $s$ as its input. Stacked residual blocks with convolutional layers are utilized to extract abstract features. At the top, HPNet extends to two output heads, namely policy and value. The policy head outputs a vector $P$ with three values, which represents the probabilities of selecting three possible folding actions for the next residue. The value head outputs a scalar $v$, estimating the amount of H-H contact for the whole protein sequence based on the current folding results. 

R-UCT is the search algorithm utilized in FoldingZero for promising folding positions. It incrementally grows to a search tree during the self-folding process. Each child node corresponds to one possible folding action, and stores the related statistics information, such as visit count, total reward, mean reward and prior probability. These parameters are updated during multiple rounds of Monte Carlo tree search, which consists of selection, expansion, evaluation and backpropagation. When the amount of search round reaches to the configured upper limit, next action will be selected based on these statistic information. 

R-UCT in FoldingZero does not use Monte Carlo rollouts policy in each search round, compared with standard algorithms. Because in the HP model, the search space will inflate exponentially with the increase in protein length, rollout policy will result in overwhelming computational and memory cost. As an efficient replacement, FoldingZero leverages the HPNet to expand and evaluate the unexplored leaf nodes in the search tree. The output of policy head is directly appended to the new child node as its prior probability. The value head output is utilized to update the total and mean reward values of every node that locates along the search path during the backpropagation. Heuristically guided by the HPNet, R-UCT can effectively conduct the lookahead search and node-evaluation. To ensure validity of the folding results, self-avoid restriction is applied to the R-UCT. Except this basic policy, no other heuristics or prior knowledge is utilized to augment the R-UCT. When the tree search simulation completes, R-UCT provides a normalized probability vector $\pi$ over all of the current valid actions. According to the probabilities, the agent in FoldingZero selects the most promising self-folding action for the next amino acid.  

FoldingZero repeats the above tree search process for each residual, until the whole protein sequence is traversed.

\subsection{Reinforcement learning}
To improve the quality of self-folding results, FoldingZero leverages a reinforcement learning algorithm, which is inspired by AlphaGo Zero \cite{silver2017mastering}. It is designed to improve HPNet and R-UCT iteratively in the repeated policy procedures. 

In FoldingZero, HPNet is trained in a supervised manner to match the R-UCT search results closely. Action probability $\pi$ in the R-UCT is calculated based on the raw network output $P$ and multiple rounds of tree search, so $\pi$ may be much stronger than $P$. As a policy improvement operator, $\pi$ serves as the label for the policy head of HPNet. On the other head, the amount of final H-H contact is utilized as a positive reward to evaluate the quality of the self-folding trajectory. The algorithm is designed to maximize the reward to obtain the most stable protein conformation. As a policy evaluation operator, the final reward works as the label for the value head. 

The training samples are generated during the self-folding process. At time-step $t$, the current folding state $s_t$, and its corresponding action probability $\pi_{t}$ in R-UCT can be immediately obtained. When a whole protein sequence is folded, the amount of final H-H contact is applied to each intermediate self-folding time-step as its reward $r$. 
For one protein sequence with length $L$, eventually it can generate $L-1$ training samples $(s_t, \pi_t, r)$, and we store all of them into a database. 

We keep training the HPNet until the configured iteration limit is reached. To ensure that we can always utilize the best HPNet to guide the R-UCT, we introduce a competitive mechanism. Over a test dataset, two FoldingZero agents compete with each other; one utilizes the latest HPNet parameters, and the other is based on the previous best model. If the former one wins with more folded H-H contacts, the updated HPNet will replace the previous best model to adopt in the future self-folding, and also serve as a baseline for the following agent competition. Heuristically guided and evaluated by the updated HPNet, the tree search in R-UCT may also become more powerful. By repeating this policy procedures, both HPNet and R-UCT can keep improving iteratively. 


In the next two subsections, we describe the tree search steps in R-UCT and the HPNet architecture. 
\subsubsection{R-UCT} \label{sec:r-uct}
To effectively explore the possible folding space, we propose a variant of UCT algorithm named R-UCT. In the protein HP model, every different protein sequence has a corresponding theoretical upper bound of H-H contact number. R-UCT utilizes this upper bound as a regularization for the  exploitation component. Specifically, a node $s$ in R-UCT can be reached by taking a specific folding action $a$ from its parent node. Each node stores a set of statistics values, $\left \{N(s,a), W(s,a), Q(s,a), P(s,a) \right \}$. $N(s,a)$ represents the visit count of this node. $W(s,a)$ and $Q(s,a)$ record the obtained total and mean rewards by selecting this node, and $P(s,a)$ denotes the prior probability of selecting the action. One round of the tree search simulation in R-UCT can be divided into three steps:

\paragraph{Selection} During the simulation, an action is selected by the equation (\ref{eq:action}) based on the statistics in the search tree at time step $t$,
\begin{equation} \label{eq:action}
a_{t} = \argmax_a (Q^*(s_{t},a) + U(s_{t},a))
\end{equation}
where
\begin{equation}
    Q^*(s_t,a)=\frac{Q(s_{t},a)}{R_{upper}(seq)} 
\end{equation}
\begin{equation}
U(s_t,a) = c_\alpha P(s_t,a)\frac{\sqrt{\sum_i{N(s_t,a_i)}}}{1 + N(s_t,a)}
\end{equation}
where $a \in \mathcal{A}(s)$ represents all available actions that lead to corresponding candidate nodes. In (\ref{eq:action}), the first term $Q^*(s_t,a)$ represents the exploitation component, which prefers the nodes with high folded H-H contact score. The second term $U(s_t,a)$ is the exploration component, which favors the nodes that have been relatively rarely visited. $c_\alpha$ is a hyperparameter to balance exploitation and exploration. 


According to the proof by Hart-Istrail~\cite{hart1996fast,istrail2009combinatorial}, given a protein sequence $seq$, the optimal number of H-H contacts $Opt(seq)$ in the 2D HP model exists a theoretical upper bound ${R_{upper}(seq)}$. Divided by this upper bound, $Q(s_t,a)$ is scaled to $Q^*(s_t,a)$ with the same magnitude as $U(s_t,a)$. To calculate the upper bound ${R_{upper}(seq)}$, residues are indexed by their positions in the primary sequence, using the ascending order $1, 2, 3,...n$, where $n$ is the protein length. Denoting the numbers of hydrophobic residues located at odd and even positions as $O(seq)$ and $E(seq)$, respectively, we have 
\begin{equation}
    Opt(seq) \le {R_{upper}(seq)} 
\end{equation}
such that
\begin{equation}
{R_{upper}(seq)} = 2\times \min\{O(seq),E(seq)\}
\end{equation}

\paragraph{Expansion and evaluation} When reaching a leaf node $s_L$, the HPNet is utilized to evaluate its state and output the estimated reward $v_L$ and prior probability vector $P_L$. Then, the leaf node can be expanded to the search tree and its valid child node $s_i$ is initialized to $N(s_i,a_i)=0$, $W(s_i,a_i)=0$, $Q(s_i,a_i)=0$ and $P(s_i,a_i)=p_i, p_i \in P_L$.

\paragraph{Backpropagation} The statistics stored in nodes are updated backward after each simulation round. The visit count is accumulated with $N(s_t,a_t)=N(s_t,a_t)+1$. The total reward $W(s_t,a_t)$ and mean reward $Q(s_t,a_t)$ are also updated by the equation (\ref{eq:W-update}) and (\ref{eq:Q-update}). 
\begin{align}
W(s_t,a_t)&=W(s_t,a_t)+v_t \label{eq:W-update}\\
Q(s_t,a_t)&=\frac{W(s_t,a_t)}{N(s_t,a_t)}\label{eq:Q-update}
\end{align}

Self-folding probability $\pi(a_i|s) = \frac{N(s,a_i)}{\sum_{j} N(s,a_j)}$ is returned, when all the simulation rounds end. Based on $\pi$, the folding agent will select the most promising action to self-fold the residue.

\begin{figure}[!b]
\vspace{-0.5em}
\centering
\includegraphics[scale=0.25]{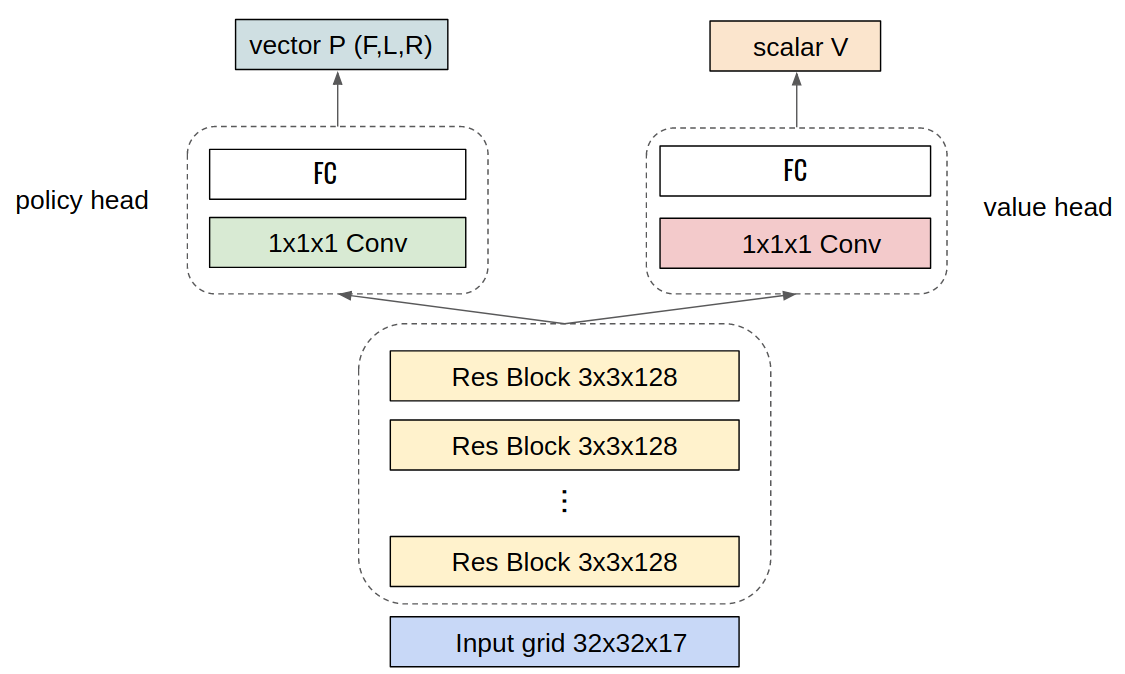}
\caption{\textbf{HPNet architecture} The output vector (F,L,R) of policy head represents forward, left and right direction based on current position. A scalar $v$ is the output of value head to estimate the reward.} 
\label{fig:network} 
\end{figure}

\subsubsection{HPNet architecture} 
The input to the neural network is defined as $X_t$, a $N\times N\times M$ image stack with grid size $N \times N$ and the number of binary channels $M$. The current folding state $s_t$ is represented as the concatenation of four binary value feature planes $[H_t, P_t, C_t, B_t ]$. They respectively correspond to H type residue, P type residue, "primary connect" and "H-H connect". For example, $H^g_t=1$ only if the grid point $g$ is occupied by the H type residue. To incorporate the sequence-folding information, we utilize $3$ steps of history states and stack them together with the current state. An extra feature plane, $E_t$ is used to represent the next residue type to be folded. It will be set as 1 if the residue is H type, or 0 if the residue is P type. The final $X_t$ is a concatenation of all these 17 planes with $X_t=[ s_t,s_{t-1},s_{t-2},s_{t-3}, E_t]$.



HPNet architecture is illustrated in Figure \ref{fig:network}. The latent spatial information is extracted from the raw lattice input by 20 stacked residual blocks with 3$\times$3 filters. Each residual block is comprised of two convolutional layers with ReLU activation function, two batch normalization layers, and a skip connection. At the top, the HPNet is split into two output heads, namely policy and value. The policy head outputs a vector $P$, representing the prior probability of each folding action. The value head outputs a scalar $v$, estimating the H-H contact score for the whole protein sequence. 



To train the HPNet, we use a cross-entropy loss for the policy head to maximize the similarity of the estimated prior probability $P$ to search probabilities $\pi$. A mean squared error is adopted to the value head to minimize the error between the predicted value $v$ and the self-folded reward $r$. Thus, the loss function for HPNet is given by:  
\begin{equation} \label{eq:loss}
l = (r - v)^{2} - \pi \log P + \beta \left\|\theta \right\|^{2}
\end{equation}
where $\theta$ represents weights of HPNet and $\beta$ is a hyperparameter that controls the L2 regularization to prevent overfitting.

\section{Experiments and analysis}



\subsection{Experimental setting of FoldingZero}
\paragraph{Self-folding}
We collect around 9000 non-redundant protein sequences from the public PDB dataset (\url{https://www.rcsb.org/}), in which any two proteins share less than 25\% sequence identity. FoldingZero utilizes the current best HPNet model and R-UCT to sequentially self-fold each protein sequence. A folding action is selected after 300 simulation rounds of the R-UCT. To increase the exploration spaces, a Dirichlet noise is added to the prior probabilities of the parent nodes, with $P(n,a) = (1-\epsilon)p_a + \epsilon \lambda_a$, where $\boldsymbol{\lambda} \sim Dir(0.03)$ and $\epsilon=0.25$.

\paragraph{Training}
We store the most recent 60,000 self-folding results into the memory. In every iteration, 256 results are sampled uniformly from the memory slots to train the HPNet. We use SGD (Stochastic Gradient Descent) with momentum 0.9 as the optimization approach, and set the initial learning rate to 0.001 and the weight decay to 4e-5.

\paragraph{Evaluation} 
To ensure that the updated HPNet model can generate higher quality prediction, we use 500 unseen protein sequences for evaluation. For every 2000 training steps with 32 batch-size, we save a new checkpoint and evaluate it. If it performs better than the previous best model, it will be used to self-fold and become a new baseline for competition in the next round.

\subsection{Evaluation}
After training FoldingZero in around two days, we evaluate it on the well-known 2D HP model benchmark dataset (\url{http://www.brown.edu/Research/Istrail_Lab/hp2dbenchmarks.html}).

\begin{figure}[htb]
\vspace{-1.0em}
\centering
\includegraphics[scale=0.20]{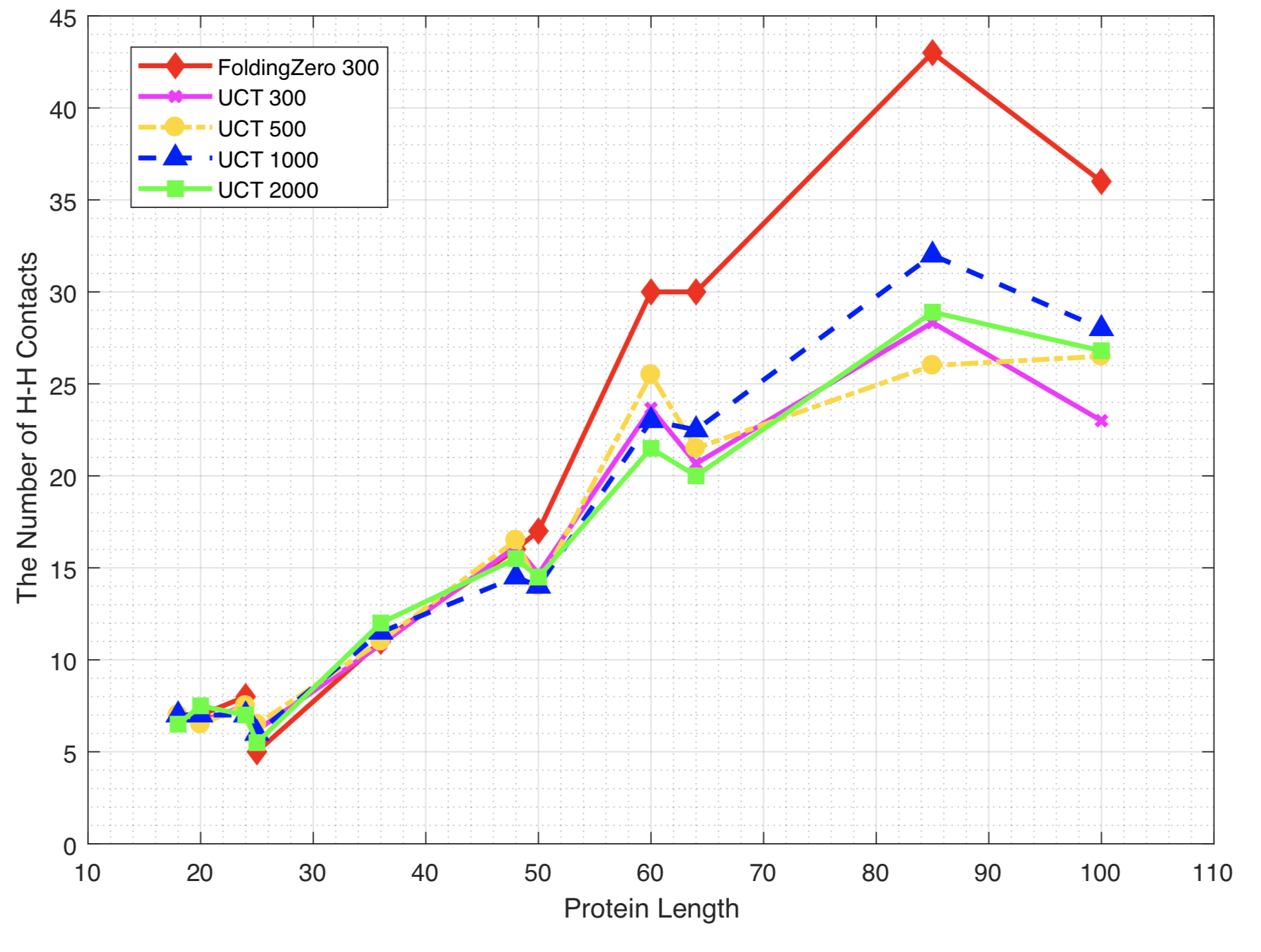}
\caption{\textbf{Comparison with UCT} evaluates the performance of FoldingZero in benchmark sequences. The plot shows total scores obtained in different simulations. For example, the red valid line denotes scores of FoldingZero in 300 simulations while the blue dashed line represent total scores from UCT in 1000 simulation rounds.} 
\label{compare_UCT} 
\end{figure}
First, we compare FoldingZero with a pure UCT based approach regarding the H-H contact score. The UCT approach employs the rollout strategy with similar information utilized by FoldingZero, except the prior probability from the HPNet. We fix the number of simulation round to 300 in FoldingZero, and adjust it in the controlled approach. As shown in Figure~\ref{compare_UCT}, with the increase in round number, the performance of the UCT algorithm slightly improves, because it can explore more state space before finalizing the selection. However, with the exponential growth of the search space, it becomes difficult to further improve performance by increasing simulation rounds. In contrast, even with much fewer simulation rounds, FoldingZero outperforms the UCT method, and the advantage is more noticeable when folding long sequences. It demonstrates that the trained HPNet can effectively guide the high-quality tree search simulations.




\begin{table*}[htb]
\centering
\vspace{-0.5em}
\caption{Free energies comparison}
\vspace{-0.5em}
\begin{tabular}{cccccc}
\hline
\multicolumn{1}{c|}{\textbf{Length}} & \textbf{FoldingZero} & \textbf{EMC} & \textbf{ENLS} & \textbf{Ant-Q} & \textbf{Optimum*} \\ \hline
\multicolumn{1}{c|}{20}                               & -9                                    & -9                            & -9                             & NA                              & -9                                 \\
\multicolumn{1}{c|}{24}                               & -8                                    & -9                            & -9                             & -9                              & -9                                 \\
\multicolumn{1}{c|}{25}                               & -7                                    & -8                            & -8                             & NA                              & -8                                 \\
\multicolumn{1}{c|}{36}                               & -13                                   & -14                           & -14                            & -13                             & -14                                \\
\multicolumn{1}{c|}{48}                               & -18                                   & -23                           & -23                            & -19                             & -23                                \\
\multicolumn{1}{c|}{50}                               & -18                                   & -21                           & -21                            & NA                              & -21                                \\
\multicolumn{1}{c|}{60}                               & -32                                   & -35                           & -36                            & NA                              & -36                                \\
\multicolumn{1}{c|}{85}                               & -49                                   & -52                           & NA                             & NA                              & -53                                \\
\hline
\multicolumn{6}{c}{*Optimum represents the opposite of the maximum H-H contact number.}                                                                                                                                              
\end{tabular}
\label{energy_compare}
\end{table*}

Second, we compare FoldingZero with other state-of-the-art heuristic approaches. A conventional metric, free energy score is utilized to measure their performance. It is defined as the opposite of the H-H contact number. EMC~\cite{liang2001evolutionary} and ENLS~\cite{guo2006exploration} were developed based on the genetic algorithm, and Ant-Q \cite{dougan2015novel} is a combined approach with evolutionary algorithm and reinforcement learning. Table~\ref{energy_compare} shows that FoldingZero achieves the comparable results and the folded free energy scores approach to the optimal ones. It is also worth noting that EMC is based on time-consuming simulation, ENLS uses memory structures to store intermediate results, and Ant-Q learns an independent Q-table for each specific sequence. Thus, when there tends to be an inordinately large number of possible solutions, the simulation rounds or memory requirements of these approaches tend to be prohibitive for longer sequences.  In contrast, the efficiency of FoldingZero does not exponentially depend on the sequence length. Even for the long sequences, it can give the decent folding results in a reasonable time period.

\begin{table*}[hbt]
\setlength{\abovecaptionskip}{1em}
\centering
\vspace{-0.5em}
\caption{Results of some representative HP sequences}
\vspace{-0.5em}
\resizebox{1.0\columnwidth}{!}{
\begin{tabular}{ccccc}
\hline
Seq ID & Length & Sequence & Optimum / Upper Bound* & FoldingZero \\
\hline
Seq1   & 20     & $(hp)_2ph(hp)_2(ph)_2hp(ph)_2$          & -9/-10                 & -9          \\
\hline
Seq2   & 20     & $h_3p(ph)_3p(ph)_3pph$         & -10/-10                & -10         \\
\hline
Seq3   & 85     & \begin{tabular}[c]{@{}c@{}} $h_4p_4h_{12}p_3(p_3h_{12})_3p(pph)_2hp_2h_2p(ph)_2$ \end{tabular}          & -53/-58                & -49         \\
\hline
Seq4   & 162    & \begin{tabular}[c]{@{}c@{}} $ph_5p_4h_4p(ph)_2(hhp)_2(ph)_2(hp)_2(phh)_2h_3p(pph)_2hp_2(h_3p_4)_2$\\$(ph)_3h_4p_2h_8(p_3h)_2h_6(phh)_2p(pph)_2h_2p_2h_2(hp)_3p_2h_4ph$\\$(pph)_2(p_4h)_2h_2p_2(ph)_2hp_3h$\end{tabular}        & NA/-78                 & -56         \\
\hline
\multicolumn{5}{c}{*Upper Bound represents the opposite of the ${R_{upper}(seq)}$ mentioned in Section 2.3.1.}                               
\end{tabular}
\label{Selected_seq}}
\end{table*}

\begin{figure}[htb]
    \setlength{\abovecaptionskip}{0.2cm}
	\centering
	\begin{subfigure}[t]{0.3\textwidth}
		\centering
		\includegraphics[scale=0.06]{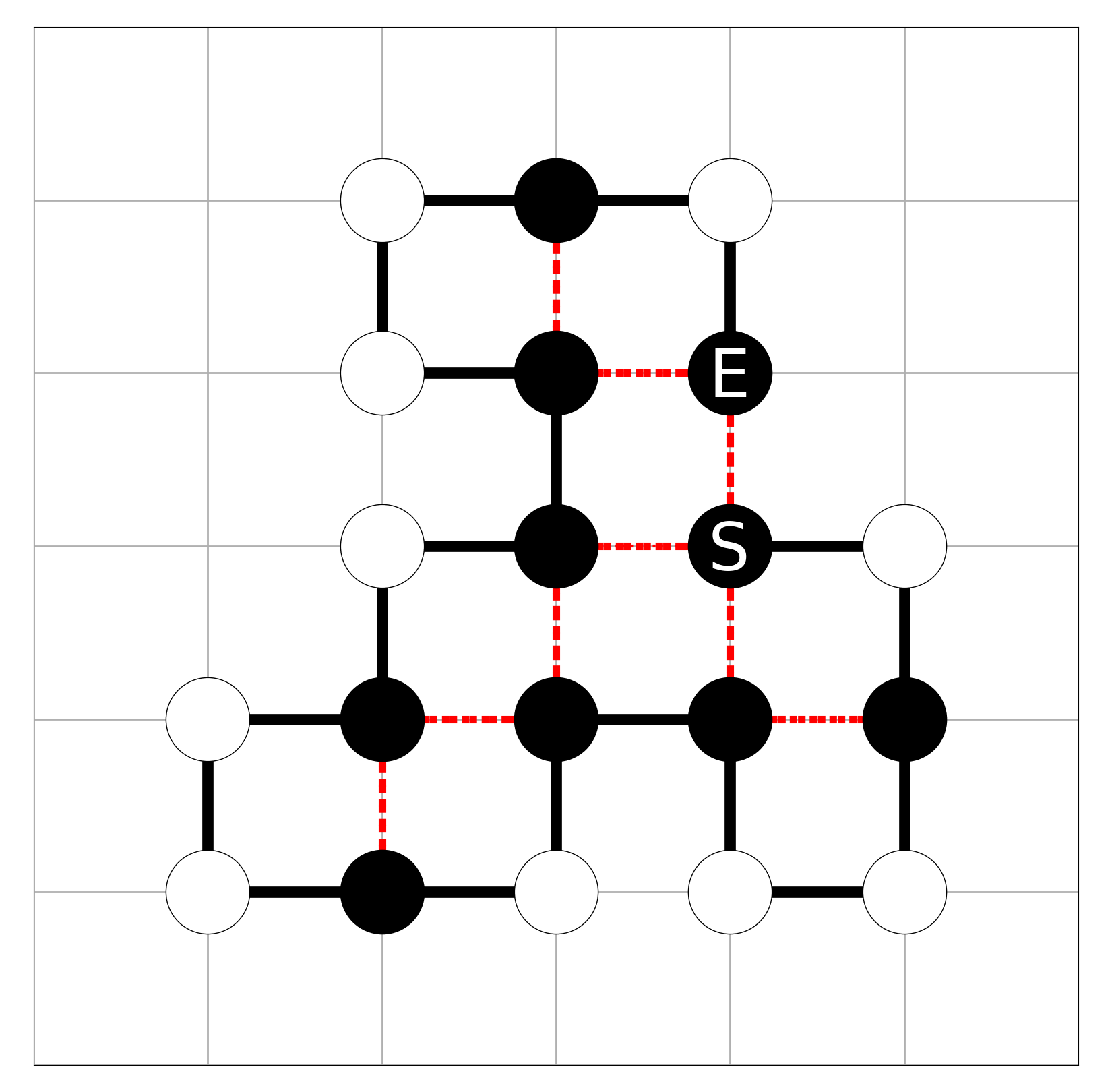}
		\caption{Seq1}\label{fig:seq1}		
	\end{subfigure}
	\begin{subfigure}[t]{0.3\textwidth}
		\centering
		\includegraphics[scale=0.06]{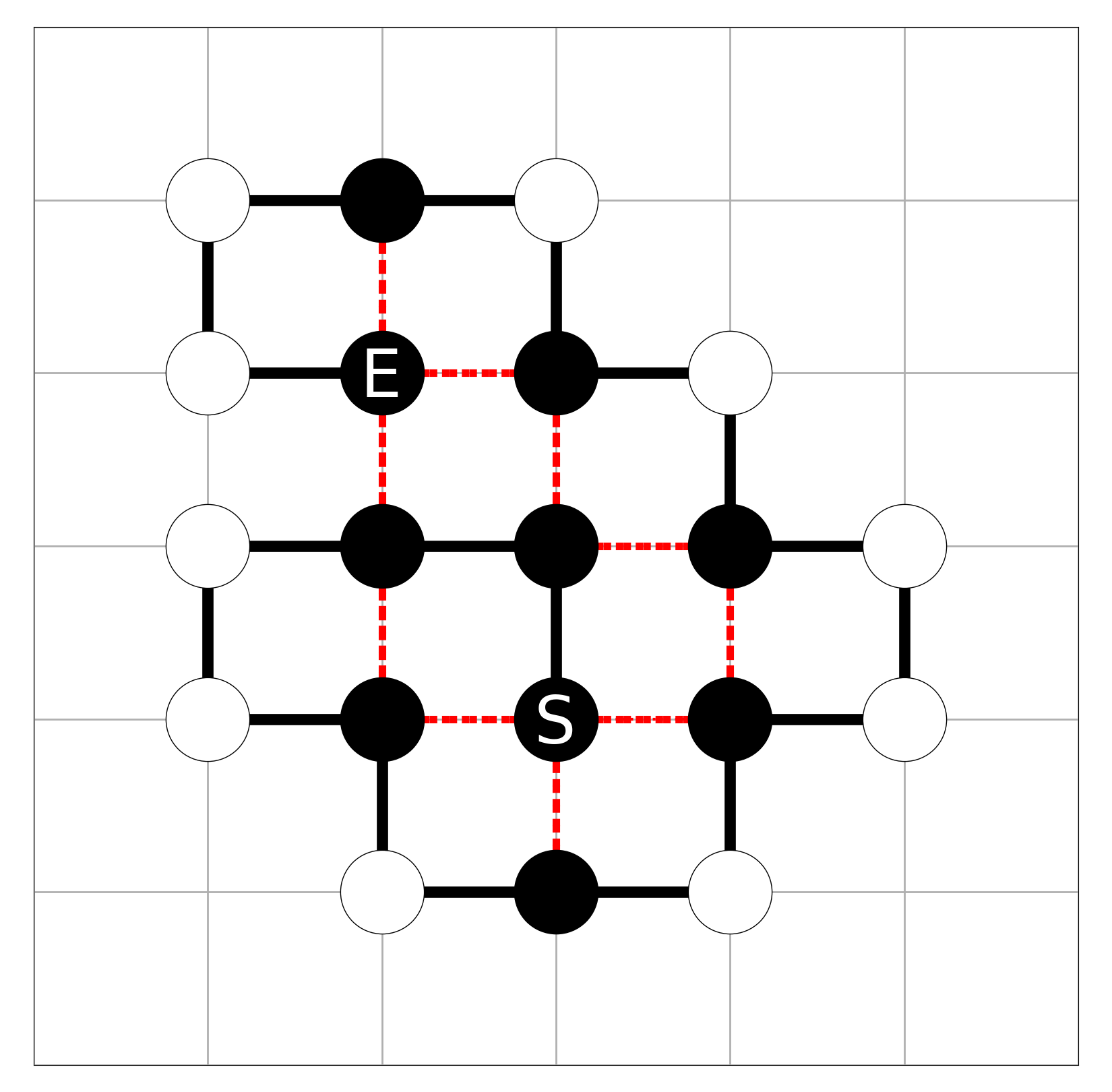}
		\caption{Seq2}\label{fig:seq2}
	\end{subfigure}
	\begin{subfigure}[t]{0.3\textwidth}
		\centering
		\includegraphics[scale=0.07]{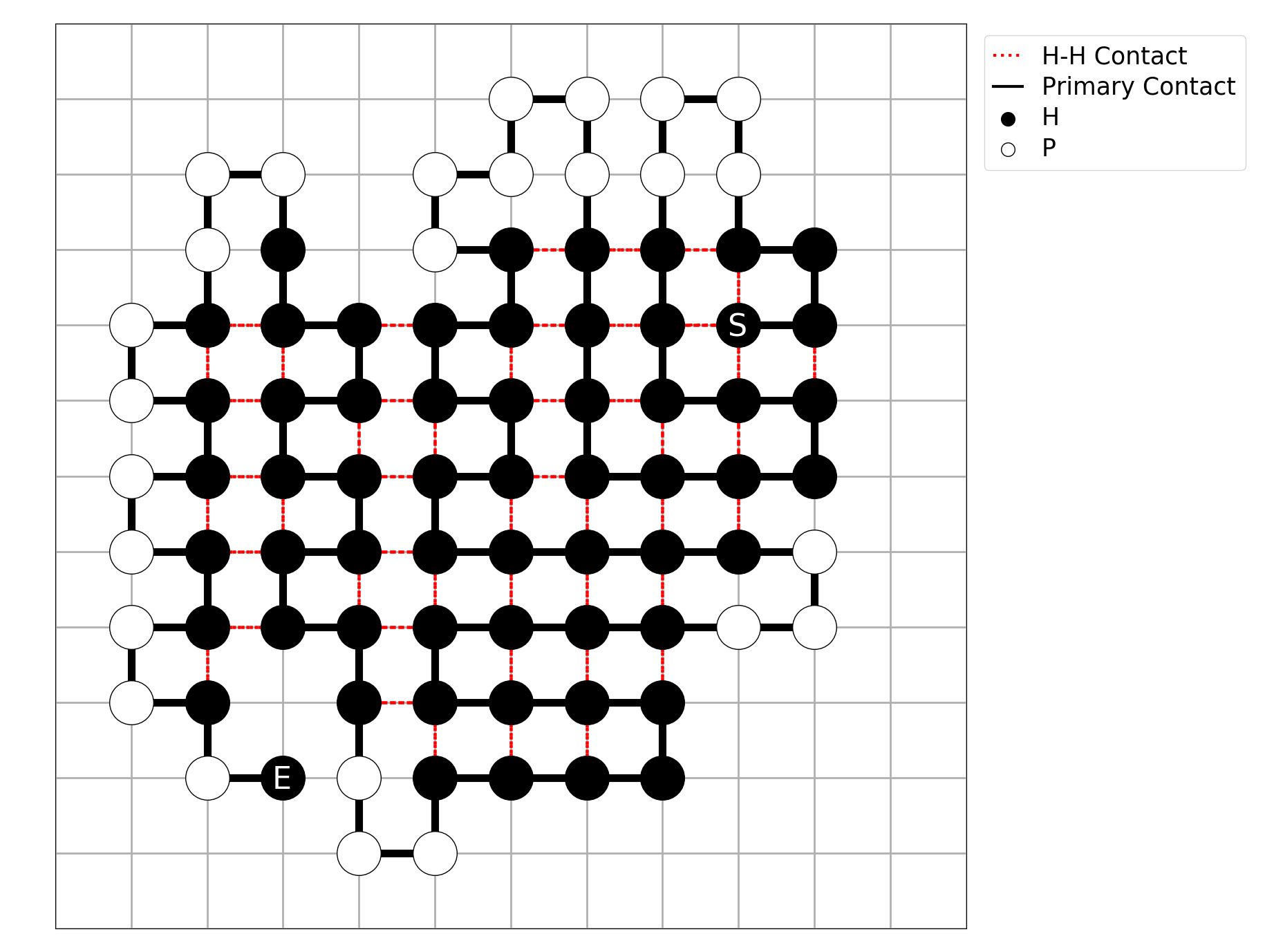}
		\caption{Seq3a}\label{fig:seq3a}
	\end{subfigure}
	\caption{Folding results of sequences listed in Table~\ref{Selected_seq}. The "S" vertex denotes the first starting residue in the sequence and "E" denotes the last ending one.}\label{fig:1}
	\label{ex_on_bm}
	\vspace{-0.3cm}
\end{figure}

\begin{figure}[htb]
	\centering
	\begin{subfigure}[t]{0.45\textwidth}
		\centering
		\includegraphics[scale=0.079]{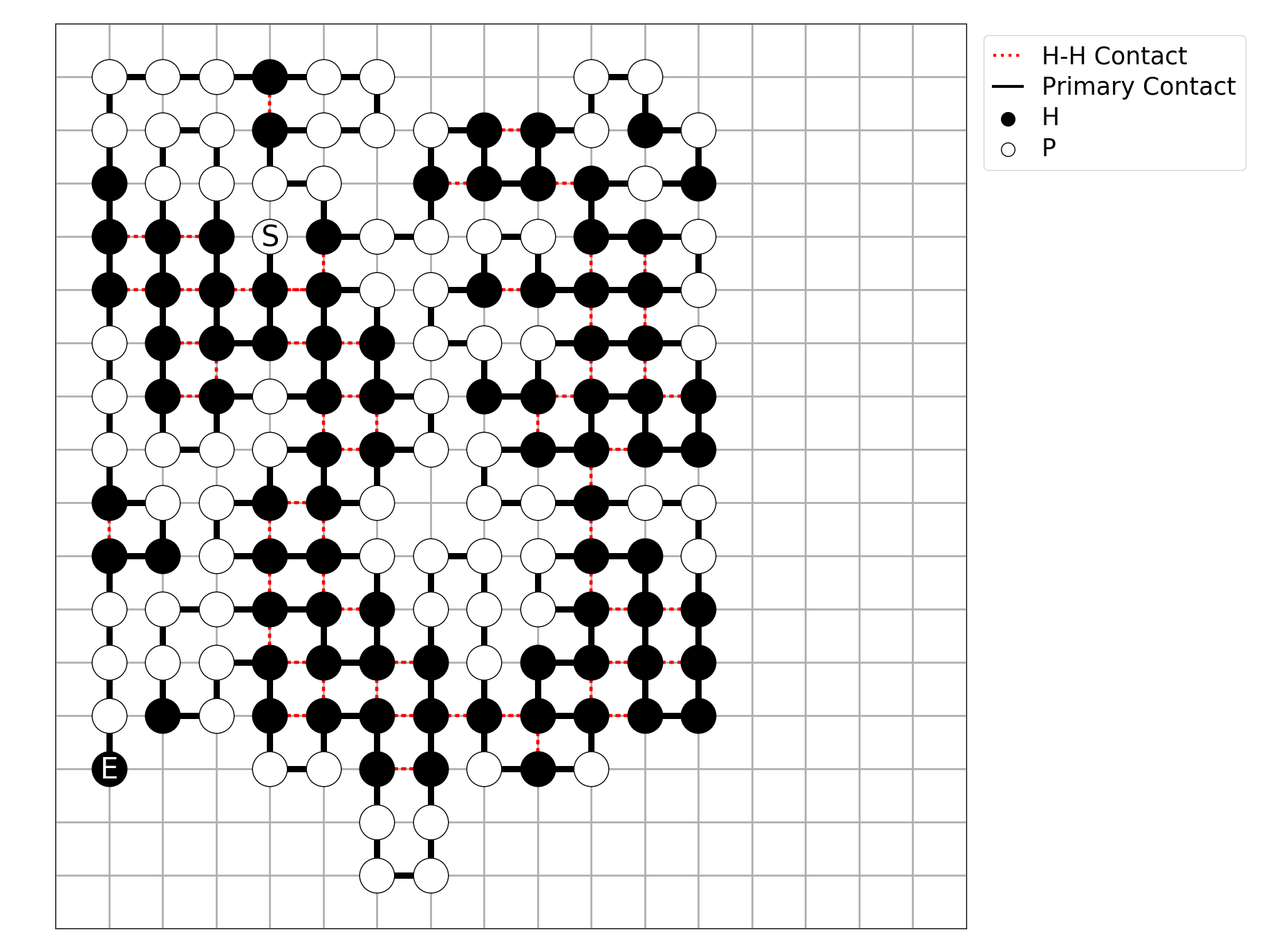}
		\caption{Seq4}\label{fig:seq4}	
		\vspace{-1cm}
	\end{subfigure}
	\begin{subfigure}[t]{0.45\textwidth}
		\centering
		\includegraphics[scale=0.079]{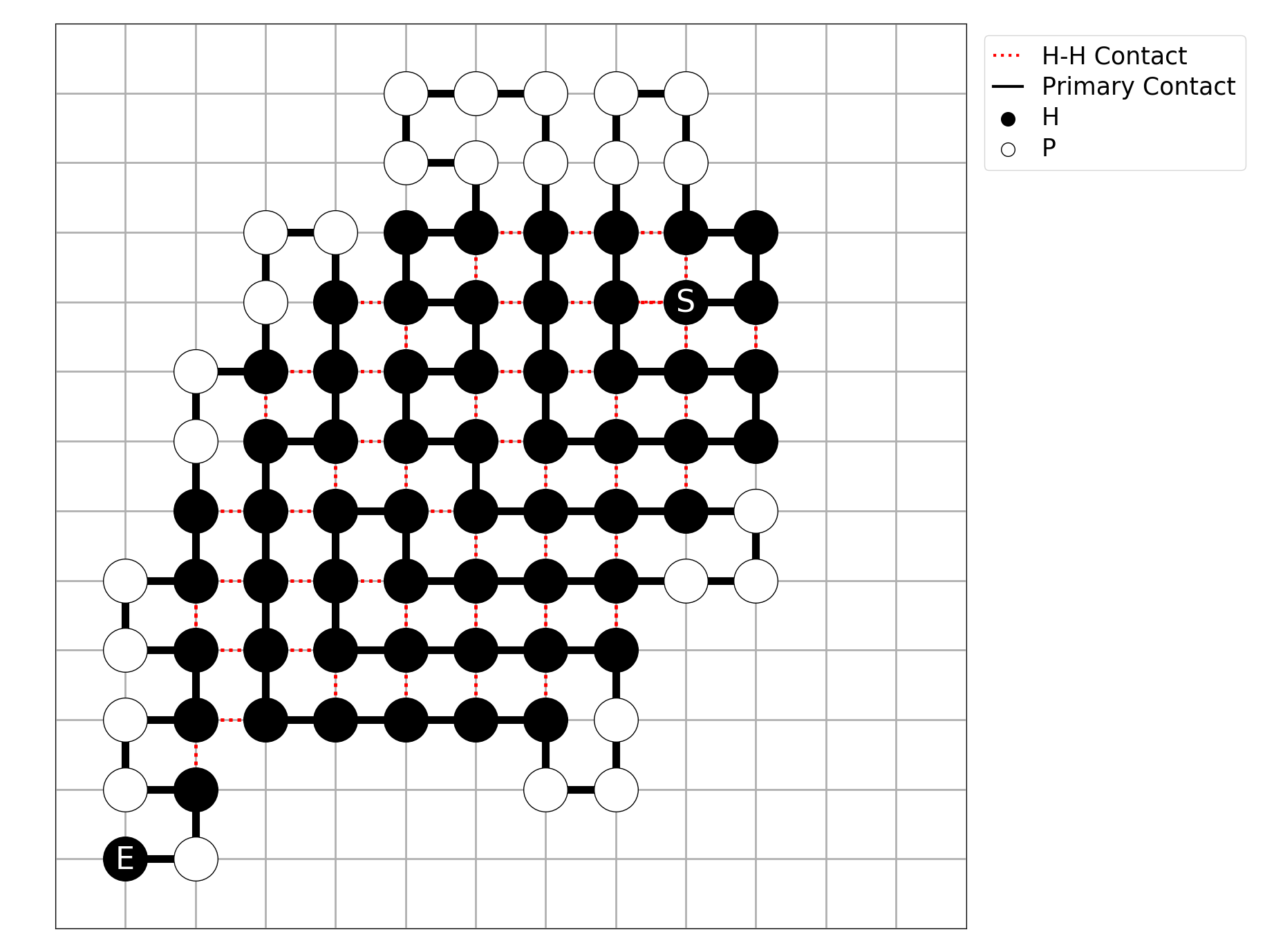}
		\caption{Seq3b}\label{fig:seq3b}
	\end{subfigure}
	\caption{\textbf{Folding results} of sequences listed in Table~\ref{Selected_seq}.}
	\vspace{-0.5cm}
	\label{85_49}
\end{figure}
\paragraph{Result analysis}
From the benchmark dataset, we select several representative protein sequences listed in Table~\ref{Selected_seq}, and visualize their folding results in Figure \ref{ex_on_bm}. We observe that FoldingZero successfully forms compact H-H cores by congregating the hydrophobic residues in the structure center and placing polar ones on the boundary. It demonstrates that FoldingZero learns the latent knowledge that hydrophobic residues are predominantly located in protein’s core, whereas polar ones are more commonly located on the surface, through DRL with extensive experiences. We also evaluate FoldingZero with some long protein sequences, which are not available in the benchmark dataset due to the limited scalability. As shown in Figure~\ref{fig:seq4}, the folded structure also exhibits the H-H core pattern. 

During the evaluation, we also notice an interesting folding result of $Seq3$, shown in Figure \ref{fig:seq3b}. For the penultimate residue of the sequence, FoldingZero still attempts to place it on the boundary, because the residue type is polar. However, this folding action causes that the last hydrophobic residue cannot form the potential H-H contact. One possible reason is that HPNet in the folding agent does not be offered the global information of the whole sequence, so that R-UCT may be misguided by the prediction. In the future work, we plan to embed the global information into the input of HPNet to further improve its capacity.

\section{Related work}
Three major types of algorithms have been developed for the HP model so far, such as approximation algorithms, combinatorial optimization algorithms and heuristic algorithms.

{\bf Approximation algorithms} offer rigorous mathematical tools and fold a protein structure within polynomial time. However, it may lead to a weak approximation ratio, resulting in a structure far from the optimal solution. Hart and Istrail proposed an approximation algorithm with ratio 3/8 of the optimal score for the 3D cubic lattice structure~\cite{hart1996fast}. An improved approximation algorithm with 2/5 performance guarantees was further developed by the same authors~\cite{hart1997lattice}. For the 2D square lattice, an approximation algorithm~\cite{newman2002new} can achieve the approximation ratio of 1/3. 


{\bf Combinatorial optimization algorithms} are exponential algorithms but can fold some protein sequences with provable optimal numbers of contacts. Backofen and Will proposed a constraint programming based approach for 3D cubic and face-centered cubic lattice structure~\cite{backofen2006constraint,mann2008cpsp}. It utilizes the hydrophobic cores concept and constraint programming technique to narrow down the solution space. However, it does not improve the worst-case time complexity, compared with a naïve search algorithm. Some other methods based on linear programming were also developed~\cite{carr2004bounding,clote2008protein}.

{\bf Heuristic algorithms} cannot guarantee the optimal solution, but they usually obtain an approximation solution in a reasonable time frame. Beutler and Dill introduced a Core-directed chain Growth method (CG) using a heuristic bias function to help assemble a hydrophobic core~\cite{beutler1996fast}. Ant colony optimization based algorithms were developed by Shmygelska~\cite{shmygelska2005ant} and Thalheim~\cite{thalheim2008protein}. Zhang et al. proposed a new Monte Carlo method, fragment regrowth via energy-guided sequential sampling~\cite{zhang2007biopolymer}. Other techniques, such as simulated annealing \cite{ullah2010hybrid}, quantum annealing \cite{perdomo2012finding}, genetic algorithms \cite{unger1993genetic} and reinforcement learning \cite{czibula2011reinforcement}, were also applied to the HP model with limited success and scalability.

\section{Conclusion}
We proposed an intelligent protein folding framework FoldingZero to self-fold the \textit{de novo} protein 2D HP structure from scratch. The HPNet and R-UCT are effectively integrated into FoldingZero to select the promising folding action. A reinforcement learning algorithm is adopted to improve the HPNet and R-UCT iteratively in repeated policy procedures. Without any supervision and domain knowledge, FoldingZero achieves comparable high-quality folding results, compared with other heuristic approaches. Without time-consuming searching and computation, FoldingZero is much more scalable and shows great potential to be applied for real-world protein properties prediction. We hope that this work could inspire future works of protein structure prediction with deep reinforcement learning techniques.

\newpage

\medskip
\small
\bibliographystyle{abbrv}
\bibliography{refs.bib}
\end{document}